\documentclass[]{achemso}

\usepackage[T1]{fontenc} % Use modern font encodings
\usepackage{amsmath}
\usepackage{xspace}
\usepackage{xcolor}
\usepackage{framed}
\usepackage{graphicx}
\usepackage{adjustbox}
\usepackage{booktabs}
\usepackage{textcomp}
\usepackage{hyperref}

\newcommand{\GImb}{G4-seq\textsubscript{IB}\xspace}
\newcommand{\Gb}{G4-seq\textsubscript{B}\xspace}
\newcommand{\GAtten}{G4-Attention\xspace}
\newcommand{\para}[1]{\paragraph{\textnormal{\textbf{#1}}.}}

\author{Shrimon Mukherjee}
\affiliation[IACS]
{School of Mathematical \& Computational Sciences, Indian Association for the Cultivation of Science, Kolkata, India}
\author{Pulakesh Pramanik}
\affiliation[IACS]
{School of Applied \& Interdisciplinary Sciences, Indian Association for the Cultivation of Science, Kolkata, India}
\author{Partha Basuchowdhuri}
\affiliation[IACS]
{School of Mathematical \& Computational Sciences, Indian Association for the Cultivation of Science, Kolkata, India}
\email{partha.basuchowdhuri@iacs.res.in}
\author{Santanu Bhattacharya}
\email{sb23in@yahoo.com, sb@iisc.ac.in}
\affiliation[IACS]
{School of Applied \& Interdisciplinary Sciences, Indian Association for the Cultivation of Science, Kolkata, India}
\alsoaffiliation[TRC]
{Technical Research Center, Indian Association for the Cultivation of Science, Kolkata, India}
\alsoaffiliation[IISC]
{Department of Organic Chemistry, Indian Institute of Science, Bangalore, India}
\alsoaffiliation[ISER]
{Department of Chemistry, Indian Institute of Science Education and Research Tirupati, Andhra Pradesh, India}

\title[G4-Attention]{G4-Attention: Deep Learning Model with Attention for predicting DNA G-Quadruplexes}

\abbreviations{G4-Attention}

\begin{document}


\begin{abstract}
  G-Quadruplexes are the four-stranded non-canonical nucleic acid secondary structures, formed by the stacking arrangement of the guanine tetramers. They are involved in a wide range of biological roles because of their exceptionally unique and distinct structural characteristics. After the completion of the human genome sequencing project, a lot of bioinformatic algorithms were introduced to predict the active G4s regions \textit{in vitro} based on the canonical G4 sequence elements, G-\textit{richness}, and G-\textit{skewness}, as well as the non-canonical sequence features. Recently, sequencing techniques like G4-seq and G4-ChIP-seq were developed to map the G4s \textit{in vitro}, and \textit{in vivo} respectively at a few hundred base resolution. Subsequently, several machine learning approaches were developed for predicting the G4 regions using the existing databases. However, their prediction models were simplistic, and the prediction accuracy was notably poor. In response, here, we propose a novel convolutional neural network with Bi-LSTM and attention layers, named G4-attention, to predict the G4 forming sequences with improved accuracy. G4-attention achieves high accuracy and attains state-of-the-art results in the G4 prediction task. Our model also predicts the G4 regions accurately in the highly class-imbalanced datasets. In addition, the developed model trained on the human genome dataset can be applied to any non-human genome DNA sequences to predict the G4 formation propensities.
\end{abstract}


\section{Introduction}
\begin{figure*}[!t]%
\centering
    {\includegraphics[width=1\textwidth]{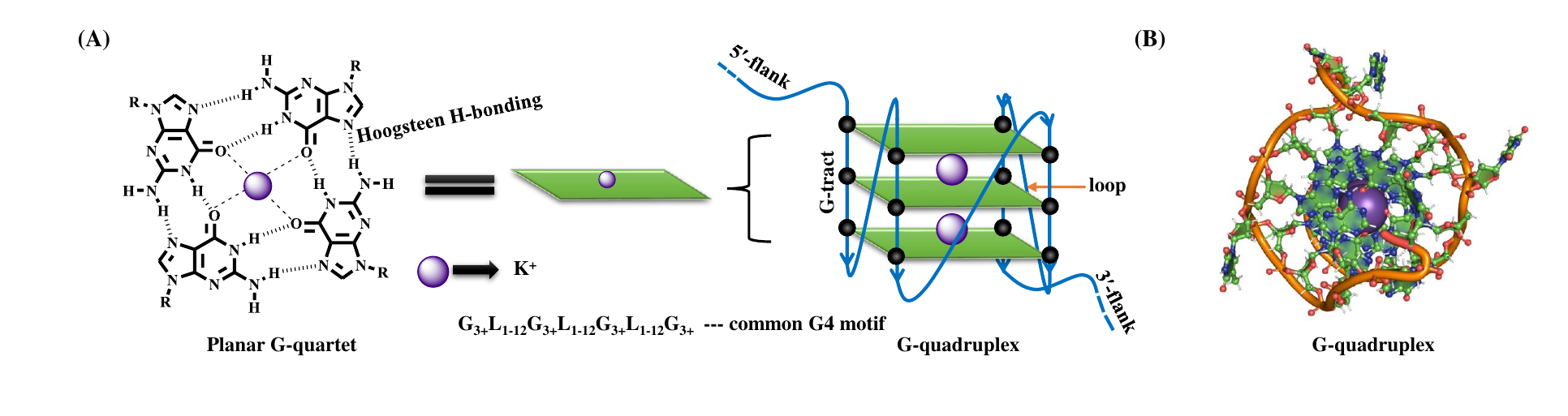}}
\caption{Schematic representation of canonical G-quadruplex structures in DNA. (A) Structure of planar G-quartet formed by four guanine bases through Hoogsteen type of H-bonding, and a central cation usually $\mathrm{K}^+$. The G-quadruplex structures are commonly comprised of three stacks planar G-quartet top to one another via $\pi-\pi$ stacking interaction. (B) Top-view of human promoter G4 found in c-MYC gene (PDB : 1XAV).}
\label{intro}
\end{figure*}

Guanine-rich DNA and RNA sequences have the ability to fold into four-stranded non-canonical secondary structures known as G-quadruplexes (G4s)~\cite{burge2006quadruplex}. These functional secondary structures were discovered in the late 80's~\cite{henderson1987telomeric} and have displayed important cellular roles in living cells~\cite{varshney2020regulation}. G4s are structurally a self-stacked arrangement of the G-quartets, a square planar organized structure formed by four guanine nucleotides connected through Hoogsteen-type of hydrogen bonding, one on top of another via $\pi$-$\pi$ stacking interaction (Fig.~\ref{intro})~\cite{chaudhuri2020recent,roy2022chemical}. These structures can form kinetically, but may be stabilized thermodynamically in the presence of monovalent cations such as $\mathrm{Na}^+$, and $\mathrm{K}^+$ under physiological conditions~\cite{jana2021thermodynamic,tucker2018stability}. Even though G4s are spontaneously formed from four consecutive tracts of three or more guanines separated by loops having various lengths under \textit{in vitro} conditions, many non-canonical G4s, that is G4s with longer loops, mismatches, and bulges, have been recently reported. It indicates that sequence patterns are less strict characteristics for the formation of G4s than previously believed~\cite{jana2021structural,hennecker2022structural}. Depending upon the physiological conditions, syn- or anti-conformation of the glycosidic bond, and intrinsic features of the sequence, G4s can adopt various topologies such as parallel, anti-parallel, and hybrid G4s~\cite{ma2020topologies}. In addition, G4s can be formed from one (unimolecular), two (bimolecular), or four (tetramolecular) distinct strands with different topologies~\cite{meier2018structure,tran2013tetramolecular}. Furthermore, the structural integrity is affected by the number of consecutive G-quartets, loop length, and the composition of nucleotides in loop regions~\cite{hao2019effects,bochman2012dna}.

The G4 motifs were significantly distributed in the telomeric regions of chromosomes, gene promoter regions, DNA replication origins, 5'-untranslated region (UTR), first intron, and exon regions of various genes~\cite{phan2002human,lago2021promoter,prorok2019involvement,chen2021translational,georgakopoulos2022alternative}. Recently, G4s were visualized using an immunofluorescence assay, which demonstrates their existence in cells and tissues~\cite{biffi2013quantitative,summers2021visualising,mulholland2016binding}. Several investigations have further revealed their potential involvement in critical biological functions such as telomeric stability, DNA replication, gene transcription, mRNA translation, and DNA repair~\cite{wu2020comprehensive,rider2022stable,jiang2021g4,de2020zuo1,jain2012dimeric,liu2023dna}. For example, in 85-90\% of cancer cells, up-regulation of telomerase enzyme was observed which blocks the activation of cell apoptosis machinery by restoring the telomeric length during the cell division process~\cite{kosiol2021g,ali2015ligand}. The terminal 3'- telomeric DNA can down-regulate the telomerase activity by forming G4s; thereby uncontrolled tumor cell proliferation is inhibited by the activation of apoptosis machinery~\cite{kosiol2021g}. Furthermore, G4s can repress the expression of various oncogenes such as MYC, KIT, Bcl-2, KRAS, HRAS, and VEGF found in cancer cells~\cite{calabrese2018chemical,paul2021g4,pandya2023g,membrino2011g4,wang2022structural,zhu2021selectivity}. G4s are therefore believed to be associated with human diseases like cancer, neurological disorders, and ageing and have emerged as a potential target for therapeutic intervention~\cite{jain2012dimeric,wang2021g,machireddy2017probing,sullivan2020molecular}.
\begin{sloppypar}
Because of the significant regulatory roles of G4s in cellular processes, numerous mathematical and computational algorithms were developed to predict the location of active putative quadruplex sequence (PQS) throughout the human genome with high accuracy~\cite{cui2023prediction}. The first PQS detection tool, quadparser was developed using the classical regular expression matching technique in C++ and Python scripting language~\cite{huppert2005prevalence}. This tool was used to detect all matching occurrences of canonical G4 motifs across the human genome by using a strict sequence pattern $G_{3-5}N_{1-7}G_{3-5}N_{1-7}G_{3-5}N_{1-7}G_{3-5}$; where G and N represent guanine, and any other nucleotide respectively. This model identified around 376,000 unimolecular PQS motifs in the human genome (\textit{hg19} reference). One of the main flaws of this tool is that it provides a yes/no binary output without considering the structural stability and \textit{in-vivo} folding propensity. Since 2005, methods that were developed on the PQS motifs, which fit the regular expression mentioned above, have been used in the majority of the research studies to predict the G4 propensity~\cite{law2010atr,gray2014g,kumari2007rna}.
More recently, a similar kind of regular expression matching tool, AllQuads has been developed to identify the inter-molecular G4s~\cite{kudlicki2016g}. Meanwhile, several in-vitro and in-vivo experiments proved the existence of non-canonical i.e., imperfect G4s along with canonical G4s in the human genome~\cite{varizhuk2017expanding}. Afterward, new methods were developed to identify imperfect G4s by incorporating the features of the newly-found imperfect G4 and by updating the algorithms incrementally. For instance, ImGQfinder~\cite{varizhuk2014improved} can detect the non-canonical intra-molecular PQS motif with a single bulge or mismatch, and pqsfinder~\cite{hon2017pqsfinder,labudova2020pqsfinder} recognizes G4s folded from G-runs with flaws such as bulges, mismatches, and G4s with longer loops. In addition, pqsfinder~\cite{hon2017pqsfinder,labudova2020pqsfinder} also gives each hit an integer score, which represents the predicted stability of the folded G4s. Further, an algorithm based on a sliding window approach, G4Hunter~\cite{bedrat2016re,brazda2019g4hunter,lacroix2019g4hunterapps} was developed to compute the quadruplex propensity score based on G-\textit{richness} and G-\textit{skewness} of a given sequence. Overall, these tools contributed significantly to the G4 exploration. However, due to the lack of vast biophysical and biochemical databases, these tools cannot perfectly predict the G4s sequences and stability.
\end{sloppypar}
Recently, G4-seq ~\cite{chambers2015high} and G4-ChIP-seq ~\cite{hansel2018genome} types of experimental methods were implemented to map G4s \textit{in vitro} and \textit{in vivo} respectively. Following that, new machine and deep learning types of artificial intelligence approaches were used to predict the G4s much more accurately using the sequencing databases. For example, Quadron~\cite{sahakyan2017machine}, a machine learning algorithm, was trained using a large database of experimental G4 regions, obtained by the G4-seq techniques in the human genome. This algorithm can detect the G4 sequences and assign a score of the detected G4s, which represents their formation propensity \textit{in vitro}. Other deep learning models, such as PENGUINN~\cite{klimentova2020penguinn}, and G4Detector~\cite{barshai2021g4detector}, a convolutional neural network (CNN) were also developed to predict the G4 regions in vitro. Since all of these models were constructed throughout the training process with the G4-seq database, new models were then going to be developed for predicting G4 sequences in cells using G4-ChIP-seq peak DNA sequences. For instance, DeepG4~\cite{rocher2021deepg4} used a CNN to predict cell-type specific active G4 regions, which is trained using a combination of G4-seq, and G4 ChIP-seq DNA sequences with chromatin accessibility measures (ATAC-seq).
Further, epiG4NN~\cite{10.1093/nargab/lqad071} was developed which uses a ResNet~\cite{he2016deep} based architecture for G4s prediction in cells. Despite the advancement of numerous machine and deep learning models using the existing databases, the prediction accuracy was quite poor and requires further improvement. In addition, all the prediction models were very simplistic. The prediction accuracy can be enhanced by incorporating additional layers, such as GRU, LSTM, attention, and dilated convolution layers~\cite{mukherjee2022deepglstm}, into the networks. Developing more sophisticated models is imperative for achieving improved prediction accuracy.\\
In this study, we proposed a novel architecture named \GAtten to predict the G4-forming sequences with improved prediction accuracy. Our work makes the following contributions:
\begin{itemize}
    \item To the best of our knowledge, we are the first ones to introduce Bi-LSTM and attention layers on top of CNN to predict the G4-forming sequences. The experimental results showed that our model achieves state-of-the-art results in the G4 sequences prediction task.
    \vspace{1.5mm}
    \item To investigate the robustness of our model, we apply it to highly class-imbalanced scenarios.
    \vspace{1.5mm}
    \item In addition, we validate \GAtten trained on human-genome dataset on non-human genome dataset.
\end{itemize}
\section{Materials and Methods}
\subsection{Datasets}
\begin{sloppypar}
In this section, we will discuss about the datasets used to train and evaluate our proposed model \GAtten.
\para{\GImb} We acquired human genome named \textit{hg19}/GRCh37 from the the UCSC Genome Browser\footnote{\url{http://genome.ucsc.edu/}}. We employed the \textit{Python} \textit{re} package to extract PQS sequences, expanding the conventional definitions of G4. This was achieved using specific regular expression patterns: $[G^{3+}L^{1-12}]^{3+}G{^{3+}}$, a standard G4 pattern with elongated loop length~\cite{guedin2010long}; $[GN^{0-1}GN^{0-1}GL^{1-3}]^{3+}GN^{0-1}GN^{0-1}G$, a enlarged G4 pattern with probable G-run breaks~\cite{mukundan2013bulges}; and $[G^{1-2}N^{1-2}]^{7+}G^{1-2}$, an uneven G4 pattern~\cite{maity2020intra}, where $L\in\{A,T,C,G\}$ and $N\in\{A,T,C\}$. We filtered the redundant and nested G4 sequences by taking into consideration at least one nucleotide. Additionally, we calculated the total number of guanines, ensuring that they are greater than or equal to 12, to facilitate the formation of three-layered G4s. This search was conducted on both strands. For each PQS, the corresponding G4 score was determined by analyzing the experimental peaks and calculating the mean of the continuous signal in the \textit{.bedgraph} format. In the instances where the PQS coordinate lacked an associated signal, the score was set as 0.0. For training purposes, G4 labels were obtained from the G4P ChIP-seq study performed on A549 cell lines, as present in the GEO database with accession number GSE133379. The top 5\% of normalized experimental scores were classified as the ``positive" class and the remaining are classified as the ``negative" class\footnote{As mentioned in this work~\cite{10.1093/nargab/lqad071}}. In this dataset most of the PQS had zero scores, therefore, the classes are highly imbalanced. Thus, we applied our model in a highly imbalanced scenario, which helps us establish the robustness of our model. The distribution of positive and negative classes is shown in Table~\ref{tab:sample_distribution}. The dataset was divided into two subparts: train and test, where the test subpart includes all the PQS belonging to chromosomes 1, 3, 5, 7, 9 and the train subpart includes all the PQS belonging to chromosomes 2, 4, 6, 8, 10-22, X, Y. The number of training and testing samples present in the dataset are 1,494,884 and 610,953.
\end{sloppypar}
\begin{table}[ht]
\centering
\caption{Distribution of Positive and Negative Samples in the \GImb Dataset}
% \vspace{-1.5em}
\label{tab:sample_distribution}
\begin{tabular}{ccc}
\toprule
\textbf{Sample Type} & \textbf{Percentage} & \textbf{Number of Samples} \\
\midrule
Positive & 5.0\% & 105,286 \\
Negative & 95.0\% & 2,000,551 \\
\bottomrule
\end{tabular}
\end{table}
\begin{table}[ht]
\centering
\caption{Dataset Statistics for \Gb dataset}
\begin{adjustbox}{width=0.75\linewidth}
\begin{tabular}{lcccc}
\toprule
 & \multicolumn{2}{c}{\textbf{$K^+$}} & \multicolumn{2}{c}{\textbf{$K^+$ + PDS}} \\
\cmidrule(lr){2-3} \cmidrule(lr){4-5}
\textbf{} & \textbf{\# Training} & \textbf{\# Testing} & \textbf{\# Training} & \textbf{\# Testing} \\
\midrule
Random     & 768,256  & 70,736   & 2,427,160 & 231,026 \\
Dishuffle  & 794,186  & 74,330   & 2,510,078 & 242,730 \\
PQ         & 784,576  & 75,457   & 1,360,065 & 131,206 \\
\bottomrule
\end{tabular}
\end{adjustbox}
% \vspace{1.5em}

\label{tab:data_distribution}
\end{table}
\para{\Gb} In the present study, we further extend the application of our model \GAtten to a balanced dataset, initially introduced in this work~\cite{barshai2021g4detector}. This dataset covers whole-genome G4 measurements across four species~\cite{barshai2021g4detector} (human, mouse, zebrafish, and drosophila)\footnote{Mouse, zebrafish and drosophila are used in the case study.}. Here, we focus exclusively on the whole-genome data about humans for training and evaluation of our model. As mentioned in this work~\cite{barshai2021g4detector}, we obtained two FASTA files for the $K^{+}$ and $K^{+}$ + PDS for the human genome, respectively and we mentioned these sets as the positive sets in G4 prediction problem. To generate the negative examples for each positive example, we followed the technique used by Barshai et. al.~\cite{barshai2021g4detector} and we constructed three datasets.
% \begin{sloppypar}
\begin{itemize}
% \end{itemize}
    \item Random: for every positive sequence, we obtained an equivalent sequence, matching in length, from a randomly selected coordinate within the human genome.
    \vspace{1.5mm}
    \item Dishuffle: for every positive sequence, we randomized its nucleotide arrangement while maintaining the frequencies of dinucleotides~\cite{jiang2008ushuffle}.
    \vspace{1.5mm}
    \item PQ: predicted G4s in the genome using the regular expression $[G^{3+}N^{1-12}]G^{3+}$, where $N$ represents any nucleotide.
\end{itemize}
For all three sets of data, we consider the genome sequences belonging to Chromosome 1 as the test sets and the remaining sequences as the training sets. The overall dataset statistics are depicted in Table~\ref{tab:data_distribution}.
\begin{figure*}[!t]%
\centering
    {\includegraphics[width=0.85\textwidth]{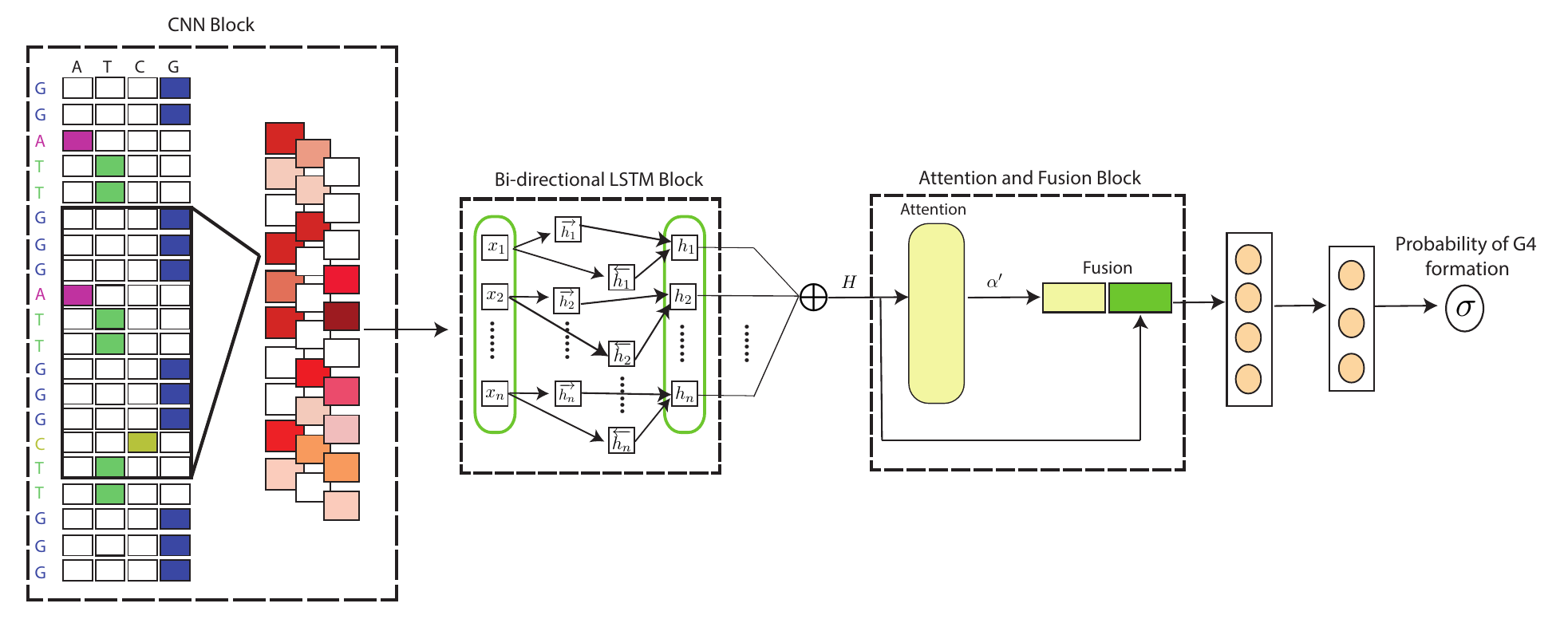}}
\caption{Schematic diagram of our proposed model \textbf{\GAtten}. Our Proposed model has three blocks: CNN block, Bi-LSTM block and finally Attention Fusion block.}\label{Gatten}
\end{figure*}
\subsection{\GAtten Neural Network Architecture}
This section starts by outlining the general idea of our proposed neural network framework namely \underline{G4} quadruplex prediction by utilizing \underline{Attention} mechanism (\textbf{\GAtten})\footnote{Our source code will be available upon acceptance of this work} for predicting whether a genome sequence forms G4 or not, followed by a detailed discussion about the different components of our proposed framework. The overview of our proposed architecture is depicted in Figure~\ref{Gatten}.
\subsubsection{Encoding of the Genome Sequence}
The nucleotide in each position of a genome sequence is represented by a 4-dimensional one-hot vector for training and testing as $A=(1,0,0,0)$, $T=(0,1,0,0)$, $C=(0,0,1,0)$ and $N = (0,0,0,0)$. We denote the one-hot vector representation for a sequence as $\mathbf{S^{O}}$ of shape $L \times 4$, where $L$ is the sequence length. $\mathbf{S^{O}}$ carries the raw information of the sequence.
\subsubsection{Feature extraction using CNN}
We automatically extract local features of $\mathbf{S^{O}}$ using a Conv1D layer. Conv1D layer is a popular architecture in genomics that automatically learns important features from raw sequences~\cite {barshai2020identifying}. More specifically, the Conv1D layer utilizes $\mathbf{S^{O}}$ as the input raw information of the sequence of shape $n \times 4$ and applies a kernel ($\Theta_{c}$) of shape $K \times F$ where $K$ is the kernel size and $F$ is the number of kernels. The output of the CNN is $x_{f}$, $x_{f} = \sigma (x_{f}^{i})$, $x_{f}^{i}\in \mathcal{R}^{n \times F}$, where $\sigma$ is a non-linear function, applied  on the intermediate value $x_{f}^{i}$ obtained through Conv1D. 
% finally we apply a nonlinearity on $x_{f}$ using the equation $x_{f} = \sigma (x_{f})$, where $\sigma$ is a non-linear func.
\subsubsection{Feature extraction using Bi-directional LSTM}
We pass the learned feature representation of $\mathbf{S^{O}}$ using our Conv1D block, which is $x_{f}$ to a Bi-LSTM layer, which captures the dependencies between the characters in a sequence of length $n$. We get the output representation $H \in \mathcal{R}^{2d_l}$, where $d_l$ represents the number of output units used in each LSTM cell. We use Eq.~\ref{eq:bilstm} for the execution of the LSTM.  
\begin{equation}
    \begin{aligned}
   \overrightarrow{h_t} &= \overrightarrow{LSTM}(x_{f}, h_{n-1}) \\
   \overleftarrow{h_t} &= \overleftarrow{LSTM}(x_{f}, h_{n+1}) \\
   % \hspace{-0.2cm}
   H &= (\overrightarrow{h_n} \cdot \overleftarrow{h_n})
    \end{aligned}
    \label{eq:bilstm}
\end{equation}
\subsubsection{Attention and Fusion block of \GAtten}
% Here we simplify the attention mechanism in natural language processing~\cite{bahdanau2014neural}. The feature representation returned in the Bi-LSTM layer, $h_t$ utilizes the Eq.~\ref{eq:context} to generate the context vector $C^{\prime}$  
% \begin{equation}
%     \begin{aligned}
%         C^{\prime} &=  [\alpha_1||\alpha_2..||\alpha_t] 
%     \end{aligned}
%     \label{eq:context}
% \end{equation}
% where, $\alpha$ is the attention weight and $h_{t}$ is the output of the Bi-LSTM. Here we calculate $\alpha_i$ using Eq.~\ref{eq: score}.
% \begin{equation}
%     \begin{aligned}
%         \beta &= h_t\Theta_a + b_a
%     \end{aligned}
%     \label{eq: score}
% \end{equation}
% where $\Theta_a$ is the learnable parameter and $b_a$ is the bias. Then we feed $\beta_i$ into the softmax layer and normalize it to $\alpha_i$ using Eq.~\ref{eq:soft}.
% \begin{equation}
%     \begin{aligned}
%         \alpha &= \frac{exp(\beta)}{\sum_{i^{'}=1}^{n}exp(\beta_{i^{'}})}
%     \end{aligned}
%     \label{eq:soft}
% \end{equation}
% The fusion layer receives $C = h_{t}$ from Bi-LSTM layer and $C^{\prime}$ from the attention layer, then we calculates the feature $X_{F}$ by weighted sum as $X_{F} = C^{\prime}C$.
Here, we use a simplified form of attention mechanism popularly used in natural language processing~\cite{bahdanau2014neural}. The attention block calculates $\alpha$ for $H$ and the values of $\alpha$ quantify the importance of each nucleotide in a genome sequence. We calculate $\alpha$ using the Eq.~\ref{eq:score}.
\begin{equation}
    \begin{aligned}
        \alpha &= H\theta_{A} + b_{A}
    \end{aligned}
    \label{eq:score}
\end{equation}
where $\theta_{A}$ is the trainable weight in the attention module, $b_{A}$ is the bias. $\alpha$ is fed into the softmax layer and is normalized to $\alpha^{\prime}$ using the following Eq.~\ref{eq:norm}.
\begin{equation}
    \begin{aligned}
        \alpha^{\prime} &= \frac{e^{\alpha}}{
        \sum_{j}e^{\alpha_j}
        }
    \end{aligned}
    \label{eq:norm}
\end{equation}
The fusion block receives $H$ and $\alpha^{\prime}$ from the Bi-LSTM and the attention blocks and subsequently computes the final feature tensor $X$ using the Eq.~\ref{eq:fuse}.
\begin{equation}
    \begin{aligned}
        X = \alpha^{\prime}H
    \end{aligned}
    \label{eq:fuse}
\end{equation}
\subsubsection{Prediction Layer}
$X$ is fed into two fully connected neural networks and finally, we applied a sigmoid activation function as the task is a binary classification problem. We used the binary cross-entropy loss using the Eq.~\ref{eq:loss} to train the \textbf{\GAtten} model.
\begin{equation}
    \begin{aligned}
        L &= - \frac{1}{B}\sum_{i=1}^{B}[Y_{i}\cdot\log(Y_{i}) + (1-Y_{i})\cdot\log(1-Y_{i})]
    \end{aligned}
    \label{eq:loss}
\end{equation}
where $Y_{i} = sigmoid(X_{i})$ and $B$ is the number of data samples in one batch. Again, in \GImb, the number of positive samples is suppressed by the number of negative samples by a large margin, resulting in the network potentially being biased towards the negative samples and failing to correctly identify positive samples when Eq.~\ref{eq:loss} is considered as the loss function.
This issue is tackled by implementing balanced class weights $W_{P} = \frac{|D|}{|C|\cdot freq_{P}}$ and $W_{C} = \frac{|D|}{|C|\cdot freq_{N}}$ for positive and negative samples respectively. Here $|D|$ is the number of samples present in the dataset and $|C|$ is the total number of classes. We slightly modify Eq.~\ref{eq:loss} to introduce class weights in binary classification loss as shown in Eq.~\ref{eq:class_w}.
\begin{equation}
    \begin{aligned}
        L &= - \frac{1}{B}\sum_{i=1}^{B}[ W_{P}\cdot Y_{i}\cdot\log(Y_{i}) + W_{N} \cdot (1-Y_{i})\cdot\log(1-Y_{i})]
    \end{aligned}
    \label{eq:class_w}
\end{equation}
\subsection{Parameter Settings}
% \textcolor{red}{I will write the parameter settings in this section.}
In this work, we fixed the length of the genome sequence to 124~\cite{barshai2021g4detector} for \Gb dataset, and for \GImb we fixed the length of the genome sequence to 1000~\cite{10.1093/nargab/lqad071}. Here, we considered one 1DCNN and one Bi-LSTM layer to train our framework \GAtten. We trained it for 10 epochs using Adam~\cite{kingma2014adam} for optimization with a learning rate of 0.001. We utilized a batch size of 1024 with a seed value of 123 to train our model. We used PyTorch-Ignite\footnote{\url{https://pytorch.org/ignite/index.html}} framework for our implementation. One NVIDIA A6000 48GB GPU, and one NVIDIA A100 80GB GPU were used to train and evaluate our model \GAtten.
\section{Results}
\subsection{Evaluation Criteria}
We determine the performance of our proposed model using two widely used metrics; the area under the ROC curve (AUC) and the area under P-R curve (AUPRC) using the Eqns.~\ref{eq:auc} and~\ref{eq:PR} respectivelly.
\begin{equation}
    \begin{aligned}
        TPR &= \frac{TP}{(TP + FN)} & FPR &= \frac{FP}{(FP + TN)}
    \end{aligned}
    \label{eq:auc}
\end{equation}
\begin{equation}
    \begin{aligned}
        Prec &= \frac{TP}{(TP + FP)} & Rec &= \frac{TP}{(TP + FN)}
    \end{aligned}
    \label{eq:PR}
\end{equation}
where $TP$, $FP$ and $FN$ are true-positive, false-positive and false-negative respectively. The AUC measures the performance of the model by calculating the relationship between the false-positive rate ($FPR$) and the true-positive rate ($TPR$) at different probabilities or thresholds. However, AUC is not optimal for imbalanced classification problems~\cite{10.1093/nargab/lqad071}; therefore we additionally evaluate AUPRC to determine precision and recall of positive G4 samples in class imbalanced scenarios.
\subsection{Research Questions}
In this section, we pose a few research questions which are central to our research work.
\begin{itemize}
    \item \textbf{RQ1.} To what extent does our model accurately predict the G4 propensities of a given genome sequence?
    % \vspace{1.0mm}
    \item \textbf{RQ2.} What is the effectiveness of our model in scenarios characterized by class imbalance, specifically in situations where the quantity of positive samples is substantially lower than that of negative samples?
    % \vspace{1.0mm}
    \item \textbf{RQ3.} How does our model perform in cross-domain testing, specifically in predicting G4 formation in non-human species?
    % \vspace{1.5mm}
    % \item \textbf{RQ4.} How does our model learns key sequence features in G4 formation?
\end{itemize}
\subsection{Model Comparison in Balanced Scenerio}
In relation to RQ-1, we compare our model with the existing computational methods that are used to predict G4 sequences in the balanced dataset \Gb. Here, we make a comparison between our model and 4 state-of-the-art methods, including Quadron~\cite{sahakyan2017machine}, pqsfinder~\cite{hon2017pqsfinder,labudova2020pqsfinder}, G4hunter~\cite{bedrat2016re,brazda2019g4hunter,lacroix2019g4hunterapps} and G4Detector~\cite{barshai2021g4detector}. All methods are applied in identical experimental conditions and utilize the optimal parameters as specified in their respective research articles.
\begin{figure*}[!t]%
\centering
    {\includegraphics[width=0.85\textwidth]{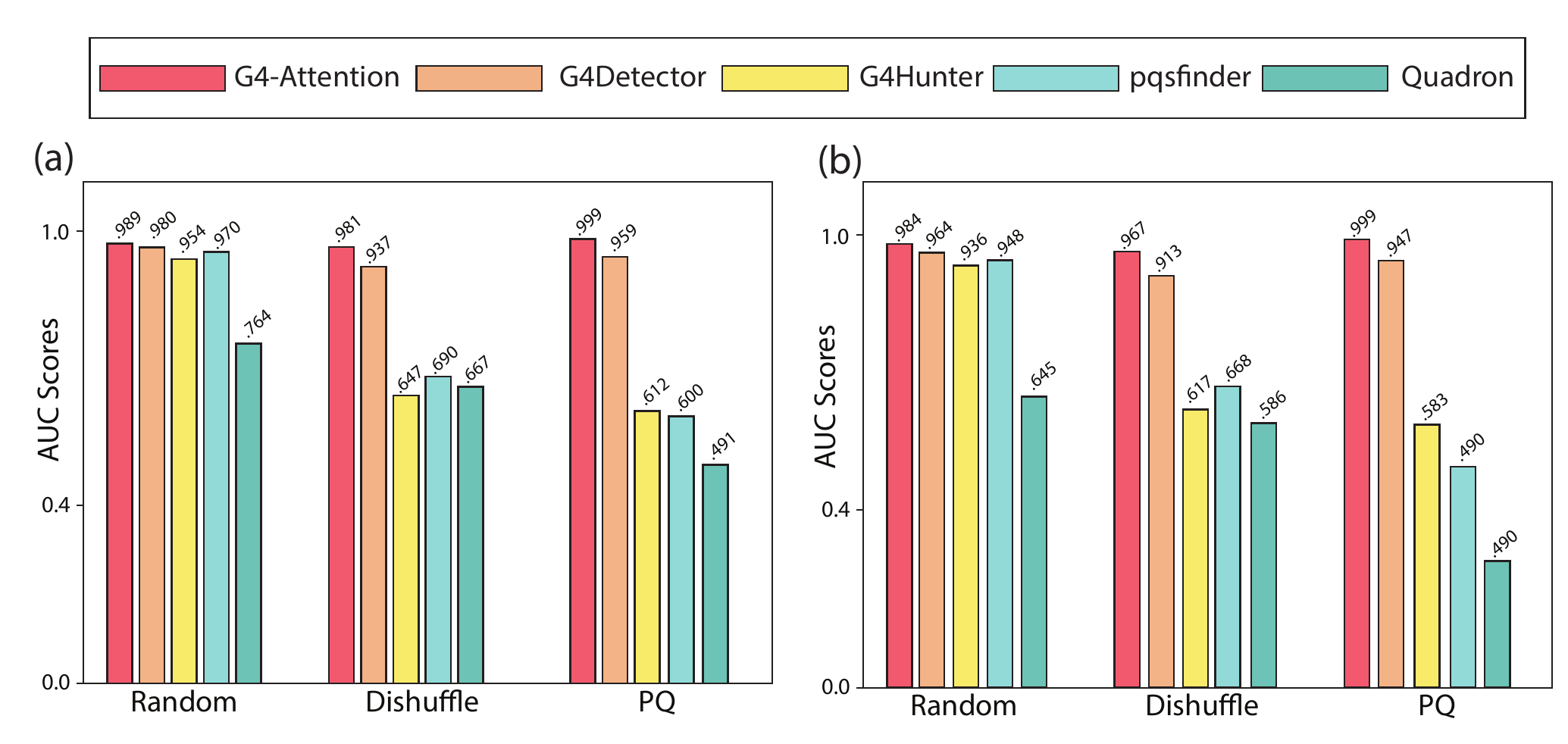}}
\caption{\GAtten outperforms all the existing techniques across all negative types (a) $K^{+}$ and (b) $K^{+}$ + PDS on held out chromosome 1 on \Gb dataset.}
\label{result_B}
\end{figure*}\\
\textbf{G4hunter}: utilizes domain knowledge-based rules, using a sliding window approach to determine G4 propensity by analyzing the GC content in a specific sequence.\\
\textbf{pqsfinder}: This method identifies G4s by locating G-tracts in the genome sequence.\\
\textbf{Quadron}: In this approach, the objective is to predict the robustness of the DNA G4 structure utilizing machine learning.\\
\textbf{G4Detector}: In this approach, the CNN-based architecture incorporates both raw sequence data and RNA secondary structure information as input.\\
\begin{figure}[!t]%
\hspace{-5.5mm}
\centering
    {\includegraphics[width=0.75\textwidth]{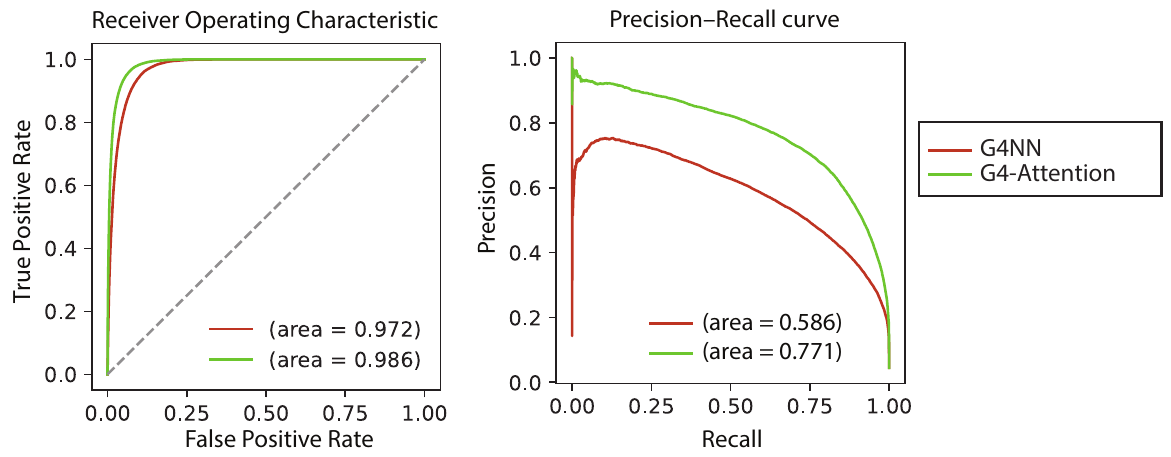}}
\caption{The performance of \GAtten on test chromosomes 1, 3, 5, 7, 9 in the \GImb dataset is depicted in this figure, where both AUROC and AUPRC are present.}
\label{result_imbalence}
\end{figure}
Fig.~\ref{result_B} (a) and (b) exhibit that our proposed architecture produces current state-of-the-art results on three negative types on two stabilizers $K^{+}$ and $K^{+}$ + PDS. The network trained solely on sequence information exhibited notable performance on the $K^{+}$ dataset, achieving AUC scores of 0.989, 0.981, and 0.999 for random, dishuffle, and PQ negatives, respectively. This performance was benchmarked against established baselines, including G4Detector, G4Hunter, pqsfinder, and Quadron. This performance was in comparison to the G4Detector method, which incorporates RNA structure probabilities into the input feature. In this study, we trained and tested \GAtten against the PQ negative set, which achieved the highest AUC score of 0.999, indicating superior performance in the most challenging of the three proposed classification problems~\cite{barshai2021g4detector}. The negative set of PQ is formulated from genomic regions characterized by a G-rich regular expression, thereby demonstrating the capability of our method to predict G4 formation in genomic segments populated with G-rich sequences~\cite{barshai2021g4detector}. In the analysis of the dataset involving $K^{+}$ + PDS stabilizers, \GAtten achieved remarkable AUC scores, registering 0.984, 0.967, and 0.999 for random, dishuffle, and PQ evaluations, respectively, which achieves current state-of-the-art on $K^+$ + PDS dataset.    
\subsection{Model Comparison in Imbalanced Scenario}
In this section, we turned our attention to RQ-2 to understand whether our proposed method \GAtten performs accurately in practical situations i.e., for the negatively skewed dataset. The occurrence of G4 structures within the human genome exhibits a pronounced negative skewness. The G4-seq experiment revealed approximately 1,400,000 G4s distributed throughout the human genome~\cite{marsico2019whole}. Given the human genome's composition of over six billion nucleotides, the frequency of G4 can be estimated to occur approximately every 4,200 to 15,000 nucleotides. This drives the necessity to assess the efficacy of \GAtten when applied to a dataset with a negative skewness. Due to the class imbalance, we measured the performance of \GAtten using the AUPRC, in addition to AUC.

We use the class imbalanced dataset named ~\GImb and measure the performance of \GAtten against the baseline model G4NN~\cite{10.1093/nargab/lqad071}, a ResNet~\cite{he2016deep} based architecture. From Table~\ref{tab:sample_distribution} we observe that the dataset \GImb consists of high negative skewness. To address this problem we used Eq.~\ref{eq:class_w} as the loss function to optimize the model parameters of the \GAtten. From Fig.~\ref{result_imbalence} we observe that it produces AUC and AUPRC scores of 0.986 and 0.771 respectively, which outperforms the existing model G4NN~\cite{10.1093/nargab/lqad071} by a large margin. This finding establishes that our model works equally well in negatively skewed datasets, thereby proving the robustness of our model.   
\begin{figure*}[!t]%
\centering
    {\includegraphics[width=0.95\textwidth]{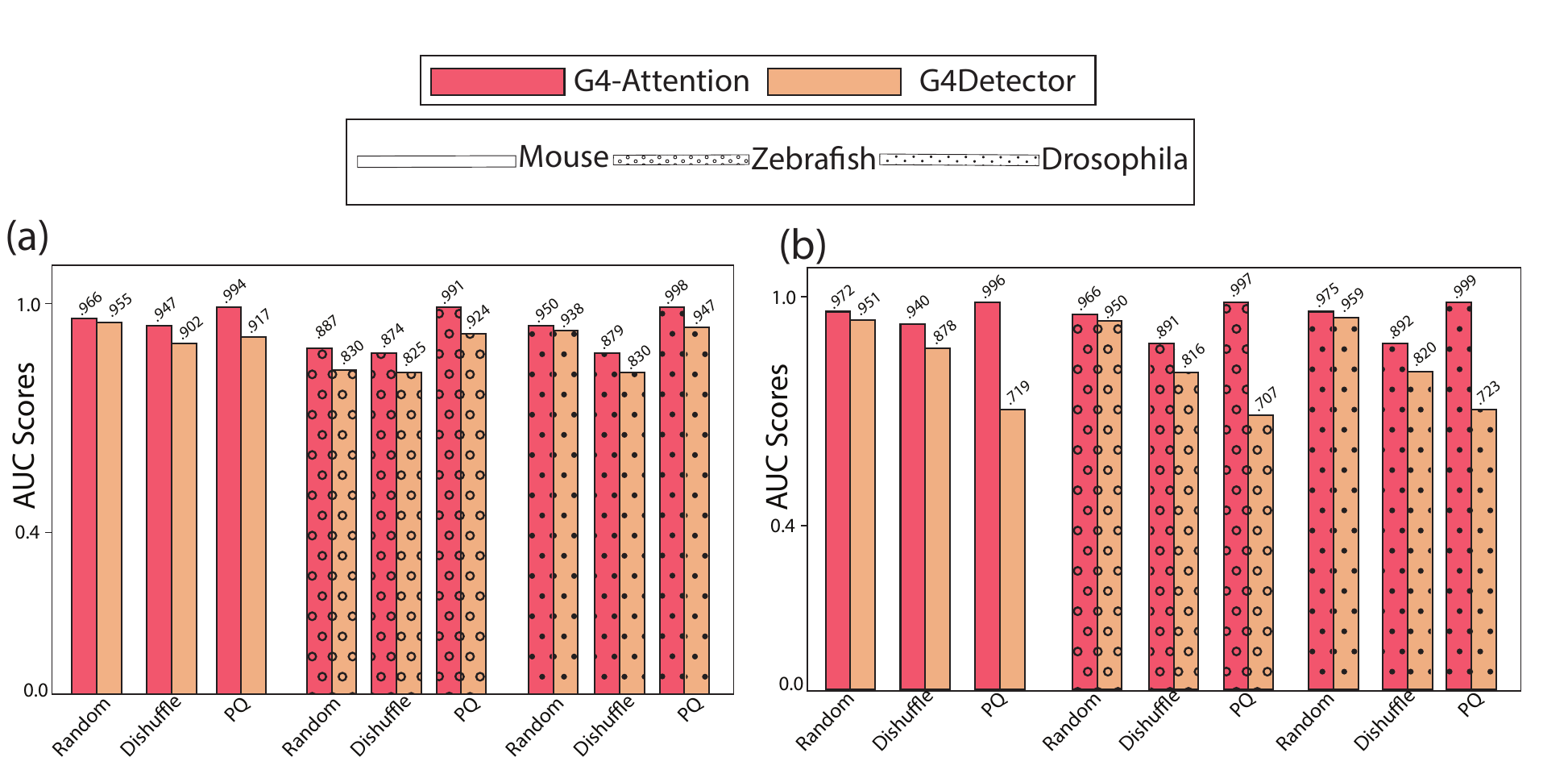}}
\caption{The comparative analysis of \GAtten's performance on novel datasets from three distinct species against G4Detector is illustrated in the figure. This evaluation specifically focuses on the AUC scores on (a) $K^{+}$ datasets and (b) $K^{+}$ + PDS datasets, including data from mouse, zebrafish, and drosophila species. AUC scores for \GAtten and G4Detector are indicated at the top of the bars.}
\label{result_Cross_domain}
\end{figure*}
\subsection{Performance of \GAtten in Non-Human Species}
To address RQ-3, we applied our model \GAtten on the multiple species dataset reported by Barshai et. al.~\cite{barshai2021g4detector}, which provides a perfect test set for \GAtten. Here, we measured the ability of our model \GAtten in a cross-domain testing scenario. We utilized \GAtten, trained on the human genome, to conduct zero-shot testing on a range of non-human species, none of which were included in the initial training dataset. From Fig.~\ref{result_B}, it is observed that \GAtten produces good results in the leave-chromosome-out test data. However, this might indicate potential overfitting of \GAtten to the \Gb dataset, or it may reveal biases specific to the human genome or human cells.

Fig~\ref{result_Cross_domain} (a) and (b) show that \GAtten produces excellent predicting performance for non-human species mouse, zebrafish, and drosophila on two stabilizers $K^{+}$ and $K^{+}$ + PDS. \GAtten predictions achieve AUC scores of 0.966, 0.947, and 0.994 on three negative datasets random, dishuffle, and PQ respectively on $K^{+}$ stabilizers. These results outperform existing baseline G4Detector results by large margins. Again, on $K^{+}$ + PDS dataset, our model produces AUC scores of 0.972, 0.940, and 0.996 on random, dishuffle, and PQ negatives respectively, outperforming the results of G4Detector. For zebrafish, our model produces AUC scores of 0.887, 0.874, and 0.991 for random, dishuffle, and PQ respectively for $K^{+}$ stabilizers, outperforming the AUC scores produced by G4Detector. Similar results are shown on $K^{+}$ + PDS for zebrafish. Finally, for drosophila, \GAtten produces AUC scores of 0.950, 0.879, and 0.998 on random, dishuffle, and PQ negatives on $K^{+}$ dataset, which surpass the results of the G4Detector. Similar results are shown on $K^{+}$ + PDS for drosophila. These findings indicate that our model \GAtten, despite being trained on the human genome, does not exhibit bias towards it. Consequently, \GAtten can be applied to the genomes of various species to accurately predict G4 formation.            
\section{Ablation Study}
\begin{figure}[!t]%
\centering
    \hspace{-2.5mm}
    {\includegraphics[width=0.75\textwidth]{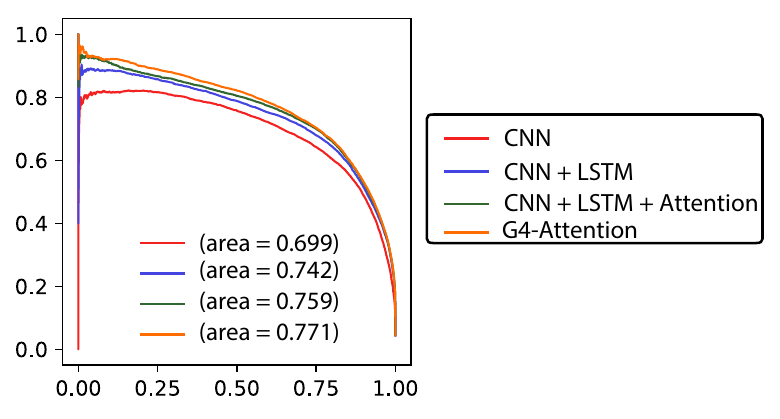}}
\caption{A series of ablation studies were performed on the class imbalance dataset \GImb. The precision-recall curves (AUPRC) have been shown.}
\label{result_ablation}
\end{figure}
Here, we perform a series of ablation experiments on class imbalanced dataset \GImb for our proposed model \GAtten. Initially, we consider CNN as our baseline model. Here, we take 1D CNN to extract features from the genome sequence and then we predict whether the genome sequence has G4 propensities or not. It achieves an AUPRC score of 0.699. Subsequently, we introduce an LSTM layer on top of the CNN layer and this model produces an AUPRC score of 0.742. Furthermore, we introduce an attention layer on the CNN + LSTM model, which increases the model performance as it produces an AUPRC score of 0.759. Finally, we implement our model \GAtten by replacing the LSTM layer in the CNN + LSTM + Attention model with a Bi-LSTM layer and this architecture produces an AUPRC score of 0.771. Fig.~\ref{result_ablation} shows that \GAtten contributes significantly towards achieving state-of-the-art results on this task. Here, we take \GImb for conducting ablation experiments because the prediction of positive G4s on the datasets with high negative skewness is a challenging task. Since the dataset is class imbalanced, we take the AUPRC metric to conduct the ablation experiments.      
\section{Conclusion}
In this work, we present a novel framework \GAtten for the task of predicting G4 formation in a genome sequence. Our model uses CNN, Bi-LSTM, and an attention-based network to predict the G4 formation of a genome sequence. We carried out our experiments on two datasets. One is a balanced dataset and another is a class imbalanced dataset with high negative skewness. Experimental results show that our model produces new state-of-the-art results on these datasets. We also carried out experiments on the genome sequences of non-human species such as mouse, zebrafish, and drosophila. Experimental results show that our model trained on the human genome not only produces excellent results on in-domain test data but also produces state-of-the-art results for out-domain test data as well. This shows that we can apply \GAtten on any genome sequence to predict its G4 propensity. These observations, collectively, lead to the overall conclusion that we have successfully developed a deep-learning model with Bi-LSTM and attention layers using human genome sequences, named \GAtten which predicts G4 DNA formation across diverse datasets, including both balanced and class-imbalanced sets, spanning various species such as human, mouse, zebrafish, drosophila, in comparison to alternatives models.
% \section{Data and Software Availability Statement}
% % \begin{sloppypar}
% The datasets and the source codes to reproduce the results present in the study are available at Zenodo Link~\url{https://zenodo.org/records/10695811}.
% \end{sloppypar}
\section{Competing interests}
The authors declare that there are no conflicts of interest.
\section{Author contributions statement}
Prof. SB and Prof. PBC conceptualized the study. SM performed experimental work. SM and PP contributed to the analysis of the results and prepared the figures. SM and PP wrote the initial draft of the manuscript. SM, PP, Prof. PBC, and Prof. SB edited the manuscript.

\begin{acknowledgement}
This study was funded by the Indian Association for the Cultivation of Science (IACS), Kolkata, India. SM thanks IACS for a research fellowship. PP thanks the Council of Scientific and Industrial Research, India for a research fellowship.
\end{acknowledgement}
\bibliography{ref} 
\end{document}